\documentclass{svproc}
\usepackage{url}
\usepackage{graphicx}
\usepackage{booktabs}

\begin{document}
\mainmatter              
\title{Water Quality Estimation Through Machine Learning Multivariate Analysis}
\titlerunning{Water Quality Estimation}  
%
\author{Marco Cardia\inst{1} \and Stefano Chessa\inst{1} Alessio Micheli\inst{1} \and Antonella Giuliana Luminare\inst{2} \and Francesca Gambineri\inst{2}}
\authorrunning{Cardia M. et al.} 
%
%
\institute{University of Pisa, Pisa (PI) 56127, Italy,\\
\email{marco.cardia@phd.unipi.it, stefano.chessa@unipi.it, alessio.micheli@unipi.it}
\and
ARCHA S.R.L., Pisa (PI) 56121, Italy \\
\email{giuliana.luminare@archa.it, francesca.gambineri@archa.it}
}

\maketitle

\begin{abstract}


The quality of water is key for the quality of agrifood sector. Water is used in agriculture for fertigation, for animal husbandry, and in the agrifood processing industry. In the context of the progressive digitalization of this sector, the automatic assessment of the quality of water is thus becoming an important asset.

In this work, we present the integration of Ultraviolet-Visible (UV-Vis) spectroscopy with Machine Learning in the context of water quality assessment aiming at ensuring water safety and the compliance of water regulation.
Furthermore, we emphasize the importance of model interpretability by employing SHapley Additive exPlanations (SHAP) to understand the contribution of absorbance at different wavelengths to the predictions.
Our approach demonstrates the potential for rapid, accurate, and interpretable assessment of key water quality parameters.


\end{abstract}

\section{Introduction}
Water quality is a critical factor in the agrifood sector, impacting everything from crop irrigation to livestock management and food processing.
Water contamination events can lead to crop losses, soil degradation, damage of aquatic ecosystems, and biodiversity loss.
Rapid detection of such events can mitigate their effects and reduce the potential damage.
As the agrifood sector undergoes digital transformation, the need for precise and efficient water quality monitoring has become fundamental.
Traditional methods of water quality assessment, while accurate, are often time-consuming, labor-intensive, have high reagent costs, and require significant manual effort \cite{Zurawska1989Determination,Motellier2000Quantitative,Thienpont1996Validation}.
Moreover, since laboratory analyses of samples are conducted offline, the results typically reflect only a small portion of the total water supply and often become available only after the water has already been consumed.
The limitations of traditional water analysis methods underline the necessity of developing reliable, cheaper, in situ, and efficient techniques for real-time assessment of water quality. 

Soft sensing refers to the combination of indirect measurements and advanced computational techniques to estimate process variables that are otherwise difficult to measure directly.
In various industrial and environmental applications, soft sensing combines data from indirect sources to provide accurate and real-time estimations of these variables.
This approach leverages a wide range of data sources and analytical methods, offering a cost-effective and scalable solution for monitoring complex systems.
Spectroscopy is one such technique that can be integrated into soft sensing.
It measures the interaction between matter and electromagnetic radiation, providing a rapid and non-destructive way to analyze samples.
Ultraviolet-Visible (UV-Vis) spectroscopy has gained importance in environmental monitoring due to its ability to provide a chemical fingerprint of samples.
This technique captures a large quantity of information from the spectral signatures of water samples.
In particular, the correlation between the absorbance of light at a specific wavelength and the concentration of a particular substance in a solution is stated by the Beer-Lambert law \cite{BeerLambertLaw}.

Machine Learning (ML) enhances the capabilities of spectroscopy in the context of soft sensing.
ML algorithms analyze the spectral data to identify complex patterns and relationships within the data.
By integrating ML with spectroscopy, we can develop robust models capable of simultaneously predicting multiple water quality parameters, thereby improving the speed of assessments \cite{asheri2018contamination,ASHERIARNON2019333}.
This approach not only reduces the need for expensive and time-consuming laboratory tests but also allows for real-time monitoring and decision-making.

This study aims to address these challenges by integrating UV-Vis spectroscopy with ML techniques to develop an automated, robust soft sensor for the estimation of key water quality parameters such as Total Organic Carbon (TOC), calcium, sodium, magnesium, conductivity, and chlorides.
Our approach involves acquiring spectral data from water samples and training an MLP model to build a soft sensor able to correlate these spectral patterns with the water quality parameters.
The innovative use of AI in this context not only enhances the speed and accuracy of water quality assessment but also opens new avenues for its application in sustainable farming and environmental monitoring.

This research presents several novel contributions compared to existing works in the literature:
\begin{itemize}
    \item Integration of Real-World Data. While many studies rely on synthetic or highly controlled datasets, this research utilizes real-world data from the Tuscany water supply network.
    \item Direct estimation of water quality indicators: there are no other works in the literature that find correlations between UV-Vis spectroscopy and chemical parameters such as calcium, sodium, magnesium, conductivity, and chlorides.
    \item Multitarget Regression Using MLP. Unlike previous studies that often focus on single-target predictions or require extensive feature engineering, this research employs a MLP for multitarget regression, allowing for the simultaneous prediction of multiple water quality indicators.
    \item Comprehensive Evaluation Metrics. The study uses a combination of Root Mean Squared Error (RMSE), coefficient of determination (R\textsuperscript{2}), and Mean Absolute Percentage Error (MAPE) to evaluate the model’s performance. This evaluation allows the understanding of both absolute and relative predictive accuracy.
    \item Emphasis on Model Interpretability. By utilizing SHAP values, this study provides clear insights into the contributions of different spectral features to the model's predictions. This offers practical implications for optimizing spectral analysis processes and improving model reliability.

\end{itemize}

By leveraging the combined strengths of spectroscopy, ML, and soft sensing, this research seeks to advance the field of water quality monitoring, providing a practical and effective solution for the agrifood sector.
This integrated approach enhances the efficiency and cost of water quality assessments, ultimately contributing to safer and more sustainable agricultural practices.

\section{Related Work}
Water quality assessment is gaining significant attention due to its crucial role in various sectors, particularly in agriculture and food processing \cite{shoushtarian2020worldwide}.
Traditional approaches rely on laboratory-based methods, which involve collecting water samples and analyzing them using various chemical and physical techniques \cite{Zurawska1989Determination,Motellier2000Quantitative,Thienpont1996Validation}.
Techniques such as titration, colorimetry, and gravimetric analysis were commonly employed to measure parameters like TOC, chlorides, pH, sodium, magnesium, dissolved oxygen, and nutrient concentrations \cite{Zurawska1989Determination,Motellier2000Quantitative,Thienpont1996Validation}.
These traditional methods, while accurate, are often time-consuming, labor-intensive, and costly due to the need for reagents and specialized equipment. 

Researchers have thus explored diverse methodologies to address the need for continuous monitoring and precise estimation of water parameters.
In particular, the introduction of spectroscopic methods marks a significant advancement in water quality analysis.
Spectroscopy is a technique used to analyze the interaction between the absorption of light and matter.
UV-Vis spectroscopy, in particular, gains importance due to its ability to provide rapid and non-destructive analysis of water samples \cite{thomas2017uv}, and it allows the analysis of water quality by capturing detailed spectral signatures of water samples.

Leveraging UV-Vis spectroscopy coupled with ML algorithms is thus as a promising approach for water quality evaluation, which is essential for constantly ensuring safe drinking water and supporting sustainable development goals \cite{alves2018use,Hossain2020,kim2016}.

Several studies explore various ML models and spectral analysis techniques for the analysis of wastewater and surface water, such as river water \cite{cardia2023machine,cardia2023multitarget,cardia2024hybrid,cardia2024wastewater,zhang2022online,Vaquet2024}.
XGBoost and Random Forest prove to be efficient in the prediction of drinking water quality \cite{Kaddoura2022Evaluation,Ozsezer2023Prediction}, while Deep Neural Network demonstrates lower error values and better performance in predicting water quality indices such as Electrical Conductivity and pH \cite{Raheja2021Prediction}.
Islam et al. demonstrates the use of UV-Vis-NIR reflectance spectroscopy combined with principal component regression (PCR) for the simultaneous estimation of various soil properties including calcium, magnesium \cite{islam2003simultaneous}.
Other researches integrate the spectroscopic data and ML to develop robust models for water quality parameters estimation \cite{cardia2024hybrid,cardia2023multitarget}.
Kim et al. proposes an optical system based on UV-Vis spectrophotometry with a multiple linear regression to estimate the TOC concentrations \cite{kim2016}.
Hossain et al. proposes a support vector machine model for the detection of disinfectant in drinking water using UV-Vis spectroscopy employing principal component analysis (PCA) \cite{Hossain2020}. Similarly, Alves et al. used UV-Vis spectrophotometry combined with PCA and MLP to determine the water quality index (WQI) as defined by Brown et al. \cite{brown1972water} in river waters \cite{alves2018use}.
However, these models are limited to a narrow set of parameters and required extensive feature engineering. Furthermore they do not provide direct estimations of water quality parameters, such as calcium, sodium, magnesium, conductivity, and chlorides \cite{Vaquet2024}.


Unlike previous studies that predominantly use synthetic or highly controlled datasets, this research utilizes real-world data from the water supply network, enhancing the robustness and applicability of our model.
By employing UV-Vis spectroscopy combined with MLP for multitarget regression, this approach allows the simultaneous prediction of different chemical parameters.
These promising results in water quality assessment demonstrate the practical advantages of deploying soft sensing technique in a real environment.

\section{Method}

The spectroscopic measurements are conducted using a Zeiss UV-Vis spectrophotometer, model CLD 600 MCS 621.
It captures absorbance spectra within the wavelength range of 210 nm to 620 nm, with a resolution of 2 nm.
The spectrometer is connected to a quartz cuvette via an optical fiber.
Quartz cuvettes are selected due to their high transparency in the UV-Vis range, ensuring minimal interference with the measurements.
This instrument is equipped with a $100 mm$ optical path length, providing an extended interaction zone for the incident light and the sample, thereby enhancing measurement sensitivity.
An attenuator of $1.4mm$ is integrated into the system to regulate the intensity of the light reaching the detector, preventing saturation and ensuring the accuracy of absorbance readings even at high concentrations.

\begin{figure}
    \centering
    \includegraphics[width=0.6\linewidth]{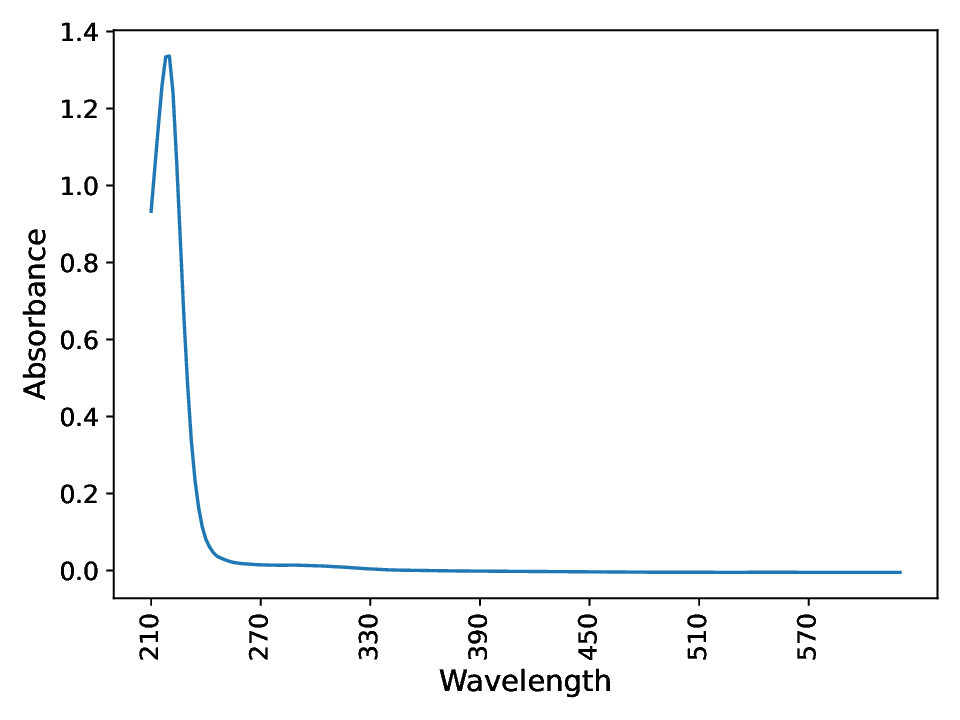}
    \caption{Absorption spectrum of the sample with the median value of the absorption spectra.}
    \label{fig:abs_spectrum}
\end{figure}

Data samples are collected from 23 different points in the Tuscany water supply network over 14 different not consecutive days distributed between February 2024 and April 2024. This approach is chosen to capture a broader range of environmental conditions within the winter and spring seasons. Furthermore, continuous daily sampling is impractical due to the time required for laboratory analysis of the collected samples.
It is important to note that these 23 sampling points are located within two cities. These locations are selected to represent key points within these two regional water supply networks, encompassing both public fountains and the water treatment plant. The inclusion of these diverse water sources, representing both raw and final (treated) water, is expected to enhance the generalizability of the model by accounting for variations in water characteristics at different stages of the water supply system.
The total number of samples collected during this period is 113.
The water samples are carefully prepared to avoid contamination and are placed in the quartz cuvette for analysis.
All measurements are carried out at room temperature, with the spectrometer calibrated using demineralized water and air before each set of readings to ensure consistency.
The corresponding absorbance spectra is the average of three measurements.
Figure \ref{fig:abs_spectrum} shows an example of the absorption spectrum of the sample having the median value of absorption spectra.

Each sample is represented as a vector of 205 real numbers, each indicating the absorbance at a different wavelength.
Since the number of samples is lower than the vector size, the data points are sparse.
To avoid the curse of dimensionality, we apply the principal component analysis (PCA) for dimensionality reduction.

For the regression task, a MLP model is employed to predict multiple water quality indicators from the UV-Vis spectra.
The model architecture consists of an input layer, one or two hidden layers, and an output layer.  The choice of using 1–2 hidden layers in the MLP model is guided by both empirical testing and literature findings in spectroscopy-based regression tasks. A single hidden layer provides a sufficient degree of non-linearity for learning spectral patterns. We have explored the addition of a second hidden layer to assess improvements in feature abstraction without significantly increasing the risk of overfitting. The hidden layers utilizes the ReLU activation function.
We have chosen this activation function due to its effectiveness in capturing non-linear relationships in spectral data.
The model is trained using a multitarget regression approach, allowing simultaneous prediction of multiple water quality indicators.
The MLP is trained for 8,000 epochs with early stopping method, the training process involves the L2 regularization to prevent overfitting.

We employ double k-fold cross-validation, comprising 4 inner folds for model selection and 5 outer folds for model assessment.
We evaluate the model's performance by using three metrics: Root Mean Squared Error (RMSE), coefficient of determination ($R^2$), and Mean Absolute Percentage Error (MAPE).
We conduct model selection by performing a random search with 128 different combinations of hyperparameters.
For each outer fold, we select the best hyperparameter values based on the performance on the validation set.
Table \ref{tab:hyperparams} lists the hyperparameters together with their ranges used for the model selection phase.
The hyperparameter ranges are selected based on prior experiments and literature findings on MLP architectures in spectroscopic analysis.
The range for the number of hidden units (350–1300) balances computational efficiency and model expressiveness. This range can be also find in literature in similar spectroscopic applications of MLPs \cite{cardia2024wastewater,cardia2023machine,cardia2023multitarget} where comparable network sizes have demonstrated effective feature extraction without excessive computational cost. Furthermore, our preliminary experiments with smaller and larger networks indicated that networks below 350 hidden units tended to underfit the data, while networks exceeding 1300 units offered diminishing returns in performance and increased training time.
The weight decay (0.001–0.002) is optimized to control overfitting. This range is selected based on both theoretical considerations of L2 regularization and empirical evidence from prior studies using weight decay in similar neural network architectures for regression tasks \cite{cardia2024wastewater,cardia2023machine,cardia2023multitarget}. Our initial experiments testing a wider range of weight decay values, from 0.0001 to 0.01, revealed that values outside the 0.001-0.002 range either resulted in insufficient regularization (leading to overfitting at lower values) or excessive regularization that hindered model learning (at higher values).

\begin{table}
    \centering
    \caption{Hyperparameters and relative range used for the random search. Values for the hyperaparameter are uniformly sampled from the reported ranges.}
    \label{tab:hyperparams}
    \begin{tabular}{lc}
    \toprule
    \textbf{Hyperparameter} & \textbf{Range} \\
    \midrule
    Hidden layers       & [1, 2] \\
    Learning rate       & [0.0001, 0.001] \\
    Units hidden layer  & [350, 1300] \\
    Weight decay        & [0.001, 0.002] \\
    \bottomrule
    \end{tabular}
\end{table}

We perform the experiments using scikit-learn version 1.2.2 and PyTorch v2.0.0 in a machine equipped with a Nvidia Quadro RTX 6000 GPU (24 GB of GPU memory).

\section{Results}
Table \ref{tab:results} summarizes the results obtained by the MLP to predict each water quality indicator in a multitarget fashion using the full vector of 205 features. For each metric, the table presents the mean value and the standard deviation across the five folds of the double k-fold cross validation.

\begin{table}
\centering
\caption{Results of multitarget regression using the MLP.}
\label{tab:results}
\begin{tabular}{lccc}   
\toprule
\textbf{Indicator} & \textbf{RMSE} & \textbf{R\textsuperscript{2}} & \textbf{MAPE} \\
\midrule
TOC & 10.63 (±4.36) & 0.91 (±0.06) & 40.77\% (±26.18\%) \\
Calcium & 12.61 (±3.07) & 0.87 (±0.05) & 12.87\% (±3.05\%) \\
Sodium & 6.59 (±1.69) & 0.89 (±0.05) & 14.26\% (±3.16\%) \\
Magnesium & 2.17 (±0.65) & 0.90 (±0.08) & 10.47\% (±2.77\%) \\
Conductivity & 75.41 (±31.66) & 0.88 (±0.10) & 7.12\% (±1.93\%) \\
Chlorides & 3.64 (±1.52) & 0.97 (±0.03) & 7.39\% (±2.64\%) \\
\bottomrule
\end{tabular}
\end{table}



In contrast, using the PCA with 15 principal components, for every water quality parameter the performance of the MLP improved, as shown in Table \ref{tab:pca_results}.

\begin{table}
\centering
\caption{Results of multitarget regression using the MLP with PCA.}
\label{tab:pca_results}
\begin{tabular}{lccc}   
\toprule
\textbf{Indicator} & \textbf{RMSE} & \textbf{R\textsuperscript{2}} & \textbf{MAPE} \\
\midrule
TOC & 7.39 (±1.93) & 0.96 (±0.02) & 35\% (±17.37\%) \\
Calcium & 11.27 (±3.03) & 0.89 (±0.05) & 11.32\% (±2.65\%) \\
Sodium & 5.68 (±1.01) & 0.92 (±0.02) & 14.80\% (±1.68\%) \\
Magnesium & 1.96 (±0.62) & 0.92 (±0.04) & 9.71\% (±3.58\%) \\
Conductivity & 73.51 (±24.37) & 0.90 (±0.07) & 7.26\% (±2.29\%) \\
Chlorides & 2.74 (±0.71) & 0.99 (±0.01) & 6.75\% (±0.72\%) \\
\bottomrule
\end{tabular}
\end{table}

\begin{figure}
    \centering
    \includegraphics[width=0.9\linewidth]{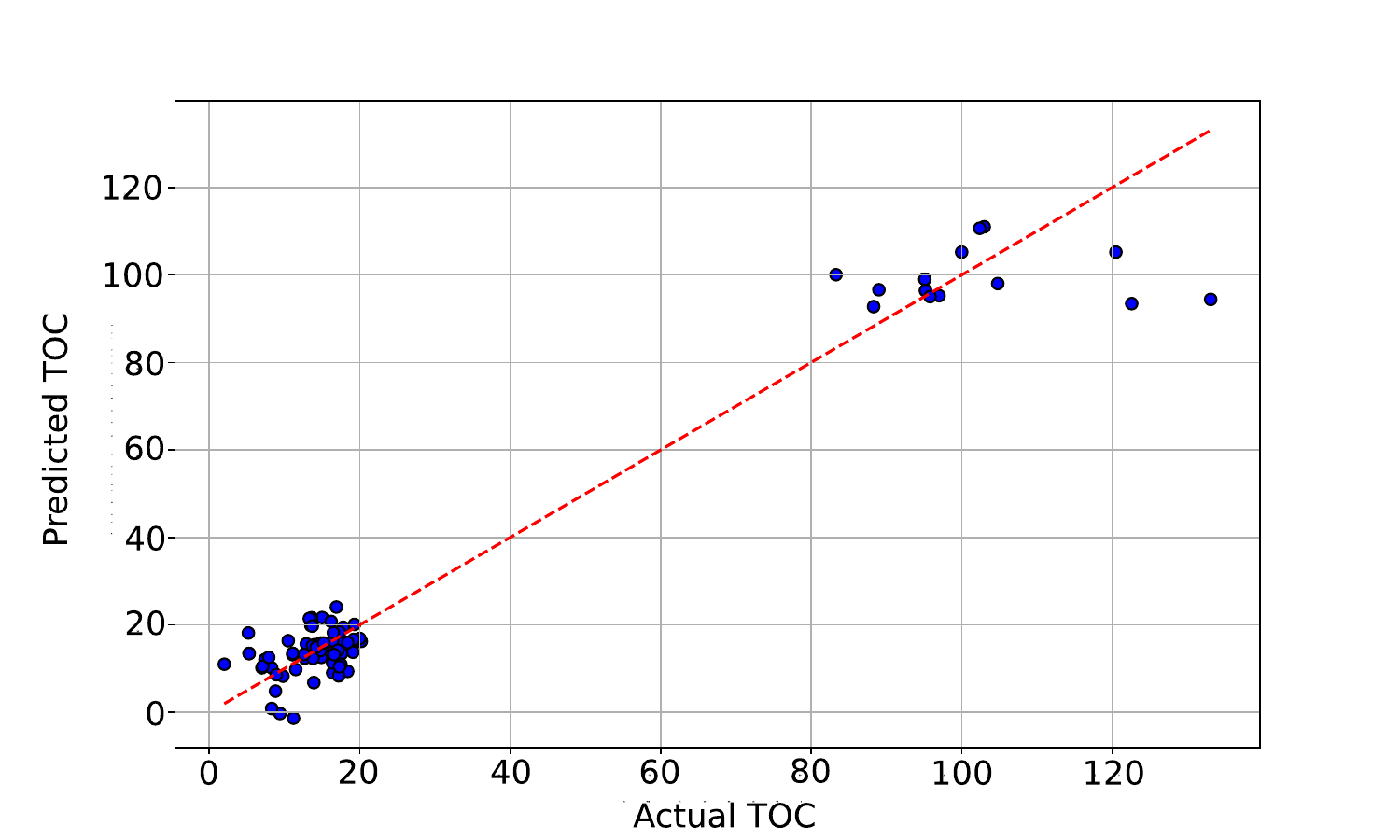}
    \caption{TOC predicted vs actual values scatter plot.}
    \label{fig:TOC_pred_vs_actual}
\end{figure}

The MLP model, when combined with PCA for dimensionality reduction, achieves an RMSE of $7.39 (\pm1.93)$, an $R^2$ of $0.96 (\pm0.02)$, and an MAPE of $35\% (\pm17.37\%)$ for TOC. These metrics indicate that while the model performs well in terms of absolute error and correlation, the relative error is relatively high. This discrepancy is attributed to the wide range of TOC values, where smaller actual values lead to higher relative errors despite low absolute errors.
It is important to note that MAPE, as a percentage-based metric, can be sensitive to the scale of the variable being predicted. For TOC, which exhibits a wide dynamic range including low values, MAPE can appear inflated even when absolute errors (in our case RMSE) are reasonable. Despite the higher MAPE, the strong $R^2$ and acceptable RMSE suggest the model captures the underlying trends in TOC effectively. Further model refinement, such as alternate feature engineering or hyperparameter tuning, could potentially reduce MAPE further, and this remains a consideration for future work. However, the current performance is considered robust for TOC prediction within the context of standard evaluation metrics in environmental modeling.
The scatter plot in Figure \ref{fig:TOC_pred_vs_actual} shows a good alignment along the diagonal, but some deviations, particularly at lower TOC values, contribute to the higher MAPE.

For calcium, the model yield an RMSE of $11.27 (\pm3.03)$, an $R^2$ of $0.89 (\pm0.05)$, and an MAPE of $11.32\% (\pm2.65\%)$. These results suggest a strong predictive capability with low relative error and a high degree of correlation between predicted and actual values.
The model's performance is consistent and reliable across the range of calcium concentrations, indicating effective feature extraction and model training.

The results for sodium show an RMSE of $5.68 (\pm1.01)$, an $R^2$ of $0.92 (\pm0.02)$, and an MAPE of $14.80\% (\pm1.68\%)$. The model demonstrates good performance and strong correlation in predicting sodium levels, with a relatively low relative error.

\begin{figure}
    \centering
    \includegraphics[width=0.9\linewidth]{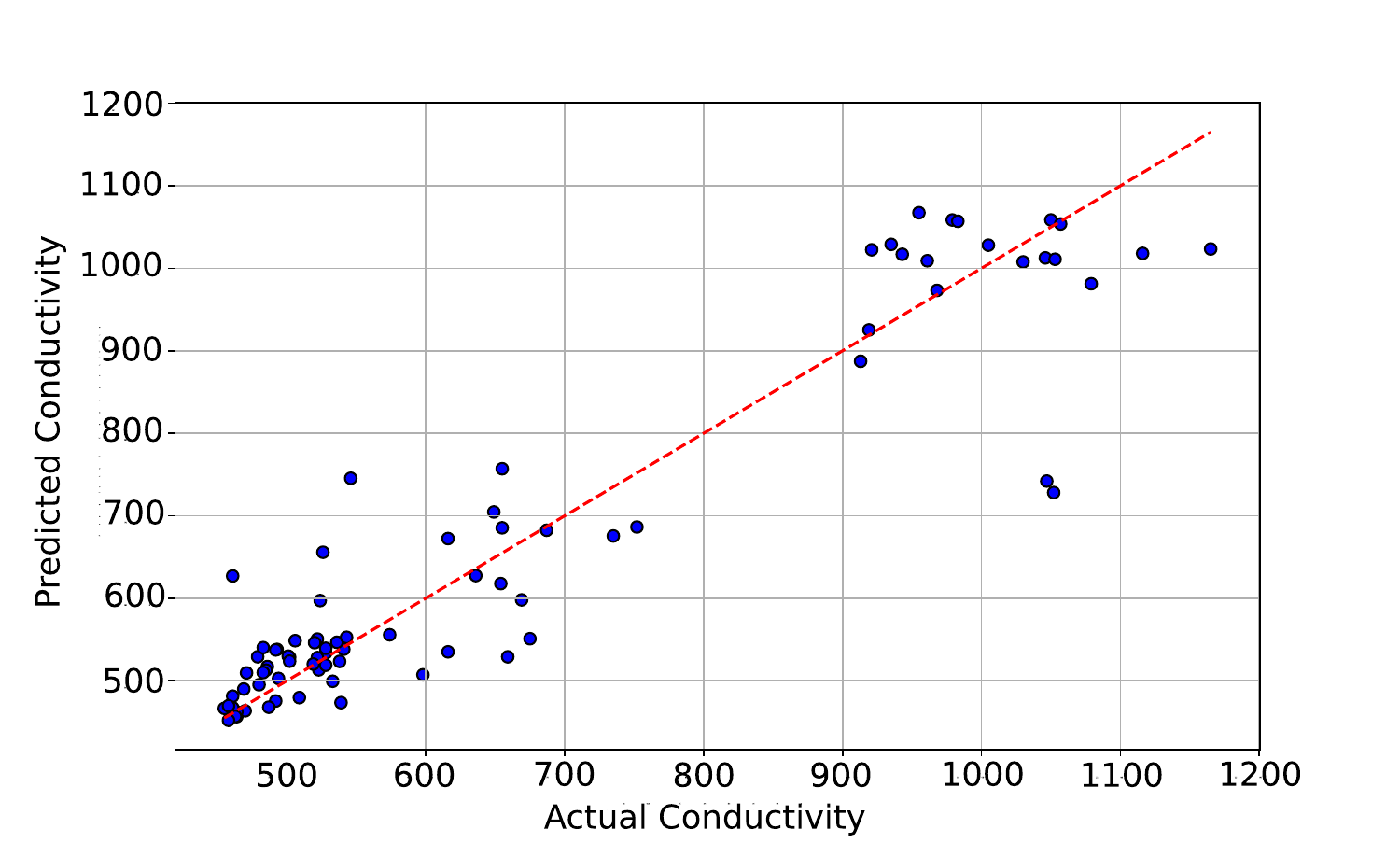}
    \caption{Conductivity predicted vs actual values scatter plot.}
    \label{fig:Cond_pred_vs_actual}
\end{figure}

Despite the higher RMSE of $73.51 (\pm24.37)$ for conductivity, the $R^2$ and MAPE values of $0.90 (\pm0.07)$ and $7.26\% (\pm2.29\%)$ respectively, indicate a strong predictive relationship with larger absolute errors.
The higher RMSE is primarily due to the larger scale of conductivity values compared to parameters like calcium and sodium, where small percentage errors result in significant absolute errors due to the magnitude of conductivity values themselves. It's crucial to understand that RMSE is an absolute error metric and is directly influenced by the scale of the predicted variable. In the context of conductivity, with its inherently larger numerical range, a higher RMSE is expected. However, the high $R^2$ and low MAPE demonstrate a strong correlation and good predictive accuracy in percentage terms.
The scatter plot in Figure \ref{fig:Cond_pred_vs_actual} shows a wider spread of points, especially at higher conductivity values.
For chlorides, the model achieves a RMSE of $2.74 (\pm0.71)$, an $R^2$ of $0.99 (\pm0.01)$ and a MAPE of $6.75\% (\pm0.72\%)$, indicating optimal predictive performance.

To better understand the performance discrepancies, additional analyses are conducted on the relative errors and residuals for each indicator. The relative error analysis highlights that the higher MAPE for TOC is driven by smaller actual values, where small absolute prediction errors translate into larger relative errors. Conversely, the high RMSE for conductivity is due to the larger scale of actual values, where even modest percentage errors result in significant absolute errors.




The MLP model demonstrates robust performance in predicting multiple water quality indicators using UV-Vis spectroscopic data. While the model performs well for most indicators, further refinement is needed to address the higher relative error for TOC and the larger absolute error for conductivity.

\subsection{Model Interpretability}

Model interpretability is a crucial aspect of ML, especially in applications such as water quality assessment where understanding the decision-making process of the model is essential.
Interpretability enhances trust, allows for validation of the model, and provides insights that can be used for further improvements.

In this study, we employed the SHapley Additive exPlanations (SHAP) values to interpret the predictions made by the MLP model \cite{NIPS2017_8a20a862}. SHAP values provide a unified measure of feature importance by considering the contribution of each feature to the prediction.
This method attributes the change in the predicted value to the presence of each feature, thus offering a clear explanation of the model's behavior.

\begin{figure}
    \centering
    \includegraphics[width=0.9\textwidth]{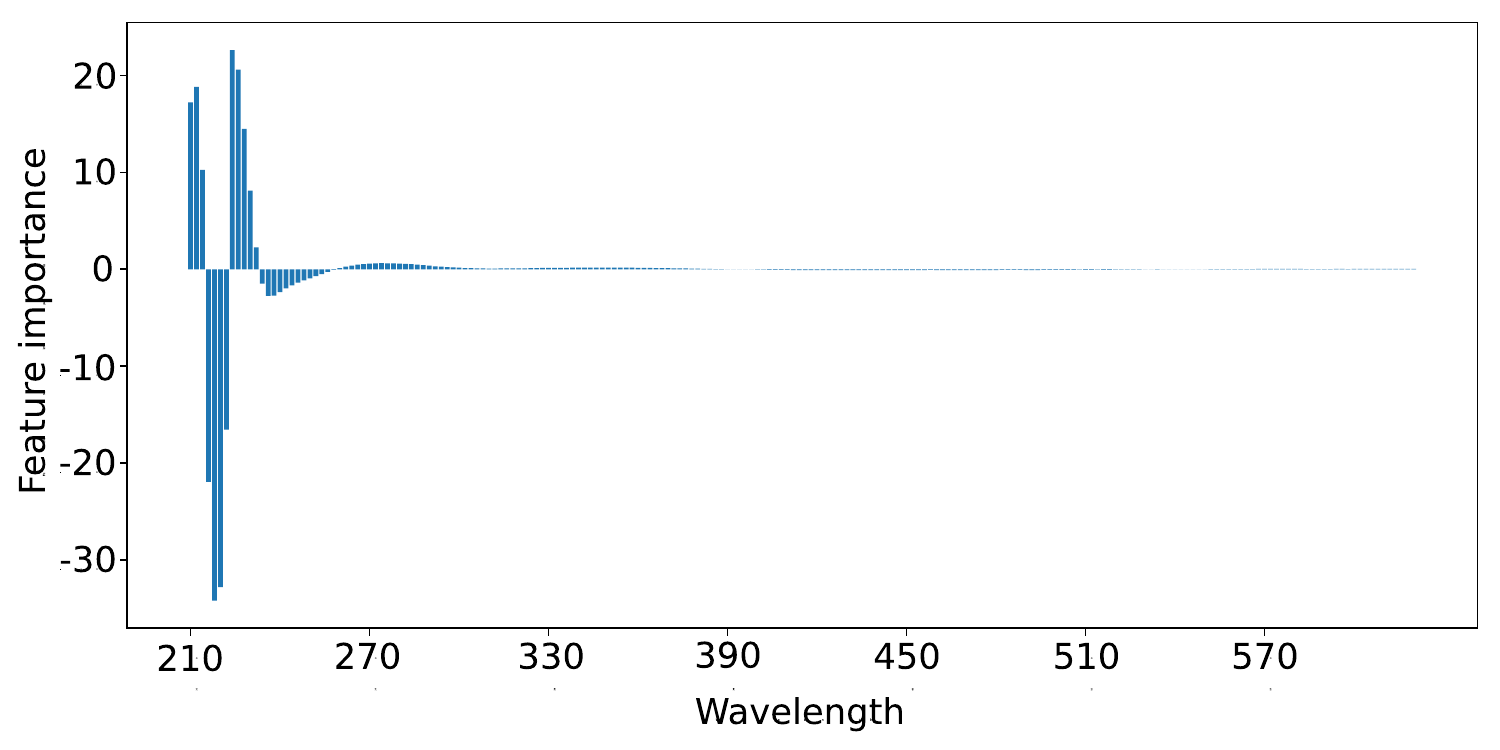}
    \caption{SHAP values for the MLP model showing the contribution of each wavelength to the prediction of TOC.}
    \label{fig:shap_values}
\end{figure}

Figure \ref{fig:shap_values} shows the SHAP values for the MLP model, highlighting the contribution of each wavelength to the prediction of TOC. Similar results were obtained for the prediction of other chemical parameters, such as calcium, sodium, magnesium, conductivity, and chlorides.
From the Figure \ref{fig:shap_values}, it is possible to observe that the wavelengths between 210 nm and 270 nm have the most significant impact on the TOC predictions. These wavelengths correspond to the UV region, where many organic compounds exhibit strong absorbance.
This indicates that the model is effectively utilizing spectral features relevant also to chemists to make accurate predictions. 
The negative SHAP values suggest that the model has learned that higher absorbance at certain wavelengths is indicative of lower TOC values, possibly due to the presence of non-organic absorbing species or other dataset-specific phenomena. 
By identifying the specific wavelengths that contribute the most to the prediction of TOC and other chemicals, it is possible to optimize the spectral analysis process.
For instance, future instruments could focus on these critical wavelengths to enhance sensitivity and reduce measurement noise.


\section{Conclusion}

This study presents a novel and effective approach for water quality assessment by integrating UV-Vis spectroscopy with ML techniques to develop a soft sensor.
By leveraging real-world data from the Tuscany water supply network, this research demonstrates the practical applicability of using multitarget regression with a MLP model for simultaneous prediction of multiple water quality indicators, including TOC, calcium, sodium, magnesium, conductivity, and chlorides.

The results indicate that the MLP model, when combined with PCA for dimensionality reduction, achieves high predictive accuracy, as evidenced by the R\textsuperscript{2} values higher than $0.88$ across all water quality indicators.

This research underscores the potential of combining spectroscopy and ML to create cost-effective, scalable solutions for real-time water quality monitoring.
Such advancements are critical for the agrifood sector, where timely and accurate water quality assessments can mitigate the risks of contamination, enhance agricultural productivity, and promote sustainable farming practices.
A valuable contribution of this study is the analysis on model interpretability. Using SHAP values, we provided insights into the contribution of different spectral features to the model's predictions. The significant influence of UV spectra on the prediction of water quality parameters aligns with established chemical understanding, confirming the relevance of our approach.

Future research directions are the extension of the study to include other spectroscopic techniques such as Near-Infrared. These techniques can provide complementary information to UV-Vis spectroscopy, potentially providing information about different water quality indicators. Expanding the dataset by collecting a broader range of samples across diverse seasons and periods would improve the model’s generalizability and robustness. Future work should aim to integrate long-term monitoring data to better capture temporal variability in water quality parameters. Additionally, transfer learning and domain adaptation techniques could be applied to adapt the developed models to different water quality datasets from various geographical regions or different types of water bodies (e.g., rivers, lakes, groundwater).
Conducting studies to monitor changes in water quality over time would further enable the analysis of temporal trends, providing valuable insights into the impact of seasonal variations, agricultural practices, and environmental policies on water quality.

\section{Acknowledgments}
This study was carried out in part within the Agritech National Research Center and received funding from the European Union Next-Generation EU (PIANO NAZIONALE DI RIPRESA E RESILIENZA (PNRR) – MISSIONE 4 COMPONENTE 2, INVESTIMENTO 1.4 – D.D. 1032 17/06/2022, CN00000022). This manuscript reflects only the authors’ views and opinions, neither the European Union nor the European Commission can be considered responsible for them.

\bibliographystyle{spmpsci}
\bibliography{mybibliography}

\end{document}